\documentclass[sigconf]{acmart}
\usepackage[utf8]{inputenc} 

\usepackage{enumitem} 
\usepackage{listings, color} 
\usepackage{hyperref} 

\definecolor{codegreen}{RGB}{34,139,34}
\definecolor{orangered}{RGB}{239,134,64}
\definecolor{darkblue}{rgb}{0.0,0.0,0.6}
\definecolor{gray}{rgb}{0.4,0.4,0.4}

\lstdefinestyle{gstcode}{
  language=sh,
  commentstyle=\itshape\bfseries\color{codegreen},
  stringstyle=\color{darkblue},
  keywordstyle=\bfseries\color{orangered},
  otherkeywords={!,.},
  deletekeywords={do,true,false},
  basicstyle=\ttfamily\small,
  breakatwhitespace=false,
  breaklines=true,
  numbersep=5pt,
}
\acmConference[ICSE 2022]{The 44th International Conference on Software Engineering}{May 21–29, 2022}{Pittsburgh, PA, USA}

\begin{document}

\title[Toward Among-Device AI from On-Device AI with Stream Pipelines]{Toward Among-Device AI \\ from On-Device AI with Stream Pipelines}

\author{MyungJoo Ham}
\authornote{The corresponding author}
\email{myungjoo.ham@samsung.com}
\affiliation{}

\author{Sangjung Woo}
\email{sangjung.woo@samsung.com}
\affiliation{}

\author{Jaeyun Jung}
\email{jy1210.jung@samsung.com}
\affiliation{}

\author{Wook Song}
\email{wook16.song@samsung.com}
\affiliation{}

\author{Gichan Jang}
\email{gichan2.jang@samsung.com}
\affiliation{}

\author{Yongjoo Ahn}
\email{yongjoo1.ahn@samsung.com}
\affiliation{}

\author{Hyoung Joo Ahn}
\email{hello.ahn@samsung.com}
\affiliation{}

\affiliation{%
  \institution{~\\\mbox{Samsung Research, Samsung Electronics}}
  \city{Seoul}
  \country{Republic of Korea}
}

\date{September 2021}

\begin{abstract}
Modern consumer electronic devices often provide intelligence services with deep neural networks.
We have started migrating the computing locations of intelligence services from cloud servers (traditional AI systems) to the corresponding devices (on-device AI systems).
On-device AI systems generally have the advantages of preserving privacy, removing network latency, and saving cloud costs.
With the emergent of on-device AI systems having relatively low computing power, the inconsistent and varying hardware resources and capabilities pose difficulties.
Authors' affiliation has started applying a stream pipeline framework, NNStreamer, for on-device AI systems, saving developmental costs and hardware resources and improving performance.
We want to expand the types of devices and applications with on-device AI services products of both the affiliation and second/third parties.
We also want to make each AI service atomic, re-deployable, and shared among connected devices of arbitrary vendors; we now have yet another requirement introduced as it always has been.
The new requirement of ``among-device AI'' includes connectivity between AI pipelines so that they may share computing resources and hardware capabilities across a wide range of devices regardless of vendors and manufacturers.
We propose extensions of the stream pipeline framework, NNStreamer, for on-device AI so that NNStreamer may provide among-device AI capability.
This work is a Linux Foundation (LF AI \& Data) open source project accepting contributions from the general public.
\end{abstract}

\maketitle

\section{Introduction}

Since the rise of deep neural networks, we have witnessed a lot of AI applications affecting our daily lives.
Such AI applications have been expanding and have started to run on devices: on-device AI~\cite{MITTechReview}.
Mobile phones run photo enhancements, video stream filtering, automatic speech recognition, noise-canceling, and object detection with deep neural networks on devices.
Televisions run video stream filtering, audio enhancements, and contents analysis.
Robotic vacuums run object detection, localization, and mapping.
Even consumer electronics products that do not appear to be ``smart'', ovens, refrigerators, and air conditioners, start to run deep neural networks on devices.
Running them on devices can preserve privacy; consumers do not want to send live video streams from robotic vacuums to servers.
It reduces network costs, especially for high-bandwidth live data such as camera streams.
It also reduces the cloud service costs; we have millions of new devices deployed each month.

The previous work~\cite{NNStreamer} is an on-device AI pipeline framework enabling developers to describe and execute AI systems as stream pipelines.
We have deployed it to recent devices of the affiliation and a few software platforms.
Implementing on-device AI systems as stream pipelines with NNStreamer has satisfied the related requirements mentioned in \cite{NNStreamer}, has improved the overall performance, and has saved the developmental costs.
After its initial deployments, we have observed new types of on-device AI applications pursued: distributed on-device AI systems where inputs, outputs, and inferences might happen on different devices.
Developers have implemented prototypes of such systems with off-the-shelf GStreamer plugins of TCP, UDP, and RTSP, allowing interconnecting independent pipelines.

We envision an open IoT ecosystem where any vendors may deploy their devices with AI services and connect to AI services of nearby devices within a house, office, or building.
Such an IoT ecosystem should be able to avoid exposing data to clouds so that AI services may fully utilize personal data safely.
Intelligence services may be executed on multiple devices in such systems, sharing hardware resources and data: ``among-device AI''.
Among-device AI may improve user experiences.
For example, devices without high computational power may provide intelligence services by connecting to nearby high-end devices, not cloud servers.
In other words, we can provide AI services with consistent quality regardless of the computation capability of the interfacing devices; thus, we can expand the interfaces and chances to provide AI services.
We may save average device costs by sharing computational power because only a few devices need high computational power.

For instance, a television without computational power for real-time pose estimation may provide an AI-based exercise trainer by connecting to a mobile phone or a home IoT hub that has enough computational power.
Conventional speakers and cameras with WiFi connections may become additional user interfaces or sensors for intelligence services.
An oven or washer/dryer unit without powerful processors may provide intelligence services of automatic speech recognition and context-aware natural language understanding by connecting to nearby refrigerators or mobile phones.

Raw network connection plugins mentioned above have been satisfactory for prototypes of the above examples.
However, they have critical issues for actual deployment, and AI application developers have expressed the need for improvement.
More critically, application developers want to connect pipelines by expressing capabilities, not IP addresses.
With initial prototypes satisfying such needs, we could have acquired further requirements from application developers and iterated a few times with a few more prototypes.
We can list up the requirements briefly as follows:

\begin{enumerate}[label=R\arabic*.]
\item Each AI or input/output service exists atomically and is deployed independently.
Such a service includes an AI inference processor, a shared input stream, or a user interface.
\item Transmission between such entities may be schemaless, or updating schema should not require recompiling or redeploying software.
It should be able to compress and synchronize transmitted data frames from different devices.
\item Discovering such entities and connecting to them should not require the knowledge of specific addresses (e.g., IP address).
Besides, there may be multiple entities discovered for a given connection request, where one of them will be connected.
\item If a connected entity becomes unavailable, it can be automatically connected to an alternative entity.
\item  Any vendors and application developers may freely use, alter, or redistribute the given software framework for any purpose without charges.
Besides, it is desirable if anyone may participate in its development to reduce fragmentation and a bias for an affiliation.
\item It should be extensible for different platforms.
In other words, it should be able to be connected with different AI pipeline frameworks (e.g., MediaPipe \cite{mediapipe} and DeepStream \cite{deepstream}) or services with different operating systems, especially those for microcontrollers (a.k.a. RTOS).
\item Each AI service is an on-device AI application that has requirements mentioned in \cite{NNStreamer}.
\end{enumerate}

Our approach extends the on-device AI pipeline platform, NNStreamer, to the among-device AI pipeline platform.
The on-device AI capability inherited from \cite{NNStreamer} R7 is preserved while new requirements (R1 to R6) are satisfied by this work.
To satisfy the new requirements, we first extend the AI stream data type (``other/tensors'' MIME) to support tensors with flexible dimensions and schemaless data streams.
Then, in addition to off-the-shelf networking plugins including TCP, UDP, and RTSP, we propose to use MQTT~\cite{MQTT} for service discovery and connections, which is applied to both publish/subscribe and workload offloading, along with synchronization methods.
We also provide a lightweight library with minimal dependencies, ``NNStreamer-Edge'', which allows implementing non-NNStreamer software packages compatible with the proposed among-device AI pipeline connections.

Our main contribution includes:
\begin{itemize}
\item Based on users' feedback, including AI algorithm researchers, application and OS developers, and device vendors, we identify requirements, propose a corresponding design, and deploy the implementation.
\item Extending the previous work~\cite{NNStreamer}, we provide mechanisms reflecting the lessons learned by deploying products.
\item We propose and deploy stream transmission protocols for among-device AI systems tested for products. The implementation is open sourced and maintained in public so that anyone may deploy products with the proposed mechanisms.
\end{itemize}

\section{Related Work}

NVidia has proposed a DeepStream AI pipeline framework based on GStreamer for NVidia hardware customers~\cite{GSTCONF18_DEEPSTREAM}.
Recent releases of DeepStream~\cite{deepstream} propose edge-to-cloud AI pipelines.
In addition to the issues of not treating tensors as 1st class citizens of stream data~\cite{NNStreamer}, it is proprietary software dedicated to NVidia hardware, which cannot meet R5, R6, and R7.

Google has proposed MediaPipe AI pipeline framework~\cite{mediapipe}, which is now targeting on-device AI applications of IoT devices in addition to cloud AI services.
Although they may have among-device AI capabilities and inter-device connectivity internally, they are not publishing documents or releasing codes of such features.

GStreamer~\cite{GStreamer} has already provided various connectivity plugins so that multimedia pipelines may listen to media servers or broadcast to media players: TCP, UDP, RTSP, and others.
It also provides data serialization plugins, GStreamer Data Protocol (GDPPAY~\cite{gdppay}), to connect remote pipelines.
Extending such capability, TRAMP~\cite{distributedGstreamerThesis} has proposed a distributed multimedia system with GStreamer pipelines partially satisfying R1, R2, and R3 for general multimedia applications, which, in turn, has given hints when we were requested to develop a software framework for edge-AI or among-device AI systems.

The previous work~\cite{NNStreamer} is an AI pipeline framework satisfying requirements for on-device AI systems of various consumer electronics.
We extend NNStreamer further with new requirements identified for among-device AI systems.

\section {Design}


As in the previous work~\cite{NNStreamer}, the first principle is to re-use well-known and battle-proven open source software components.
GStreamer~\cite{GStreamer} and its off-the-shelf plugins are still the basis of this work.
For capability-based publish/subscribe stream connections, we adopt MQTT~\cite{MQTT} via ``paho.mqtt.c'' library~\cite{paho_mqtt_c}.
We have started initial prototypes with TCP and RTSP plugins.
Then, we have applied ZeroMQ~\cite{ZMQ}, which, fortunately, has an off-the-shelf and open sourced GStreamer plugin for second prototypes.
Then, we have switched to MQTT~\cite{MQTT} for the last prototypes.
We choose MQTT because, ultimately, we want to contribute to IoT standards including Matter~\cite{Matter} and SmartThings~\cite{SmartThings}, which include MQTT.
Inter-pipeline connectivity for among-device AI is deployed as GStreamer plugins making it compatible with general GStreamer pipelines.
As a result, the provided capability is not restricted to AI applications but is available to general stream pipelines.

Initially, for earlier prototypes for internal clients, we have assumed that data serializations with GDPPAY~\cite{gdppay}, Protocol Buffers, and FlatBuffers for static tensor streams would suffice along with off-the-shelf network transport GStreamer plugins of TCP, UDP, and RTSP.
However, we have found requirements not satisfied by such design by iterating prototypes with clients.
R1, R2, R3, and R4 enforce applications to implement sophisticated network handling, which can be avoided by providing such features with NNStreamer.
Besides, implementing such features in applications inevitably leads to fragmentation and compatibility issues: every application may implement protocols differently.


\begin{figure}[t]
    \centering
    \includegraphics[width=0.98\linewidth]{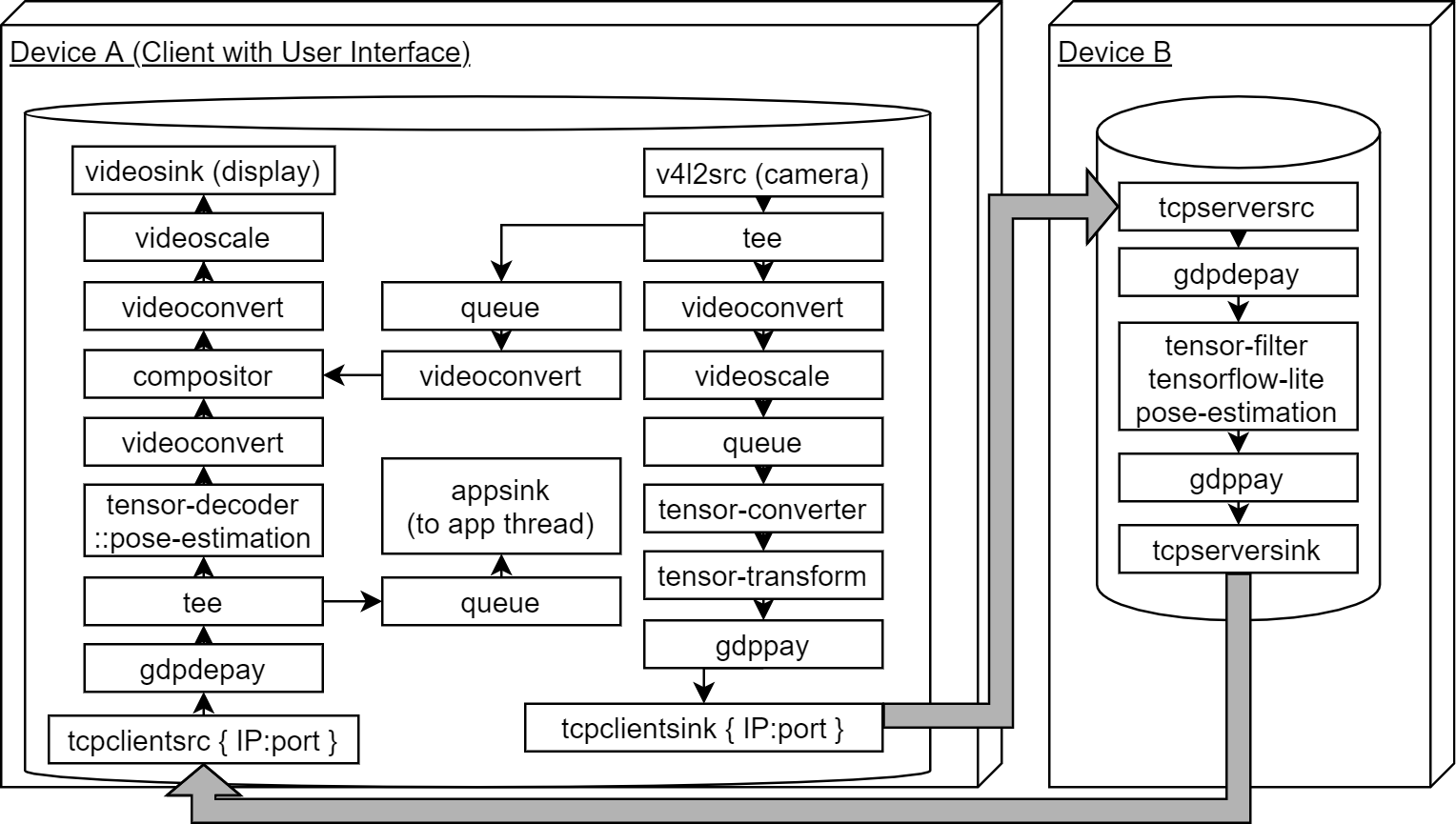}
    \caption{An offloading example with the previous work.}
    \label{fig:SIMPLEOFFLOAD_ORG}
    \Description[A pipeline offloading workload to another pipeline via TCP]{A pipeline of a client device with a user interface is offloading a pose-estimation inference to a pipeline running in another device with TCP client and server plugins. The client device has a tcpclientsink that sends requests to the server and a tcpclientsrc that receives answers from the server in its pipeline.}
\end{figure}

For example, a simple inference workload offloading (query) application with TCP plugins in Figure~\ref{fig:SIMPLEOFFLOAD_ORG} has initially appeared satisfactory.
However, having multiple clients--another additional user requirement--over-complicates pipelines.
Even with a single client, separating query (\textit{tcpclientsink}) and answer (\textit{tcpclientsrc}) filters of a client complicates pipelines with synchronization and topology issues.
Thus, this approach does not satisfy R1.
Besides, it requires clients to specify IP addresses and port numbers of servers, neglecting R3 and making it challenging to satisfy R4.

\begin{figure}[t]
    \centering
    \includegraphics[width=0.98\linewidth]{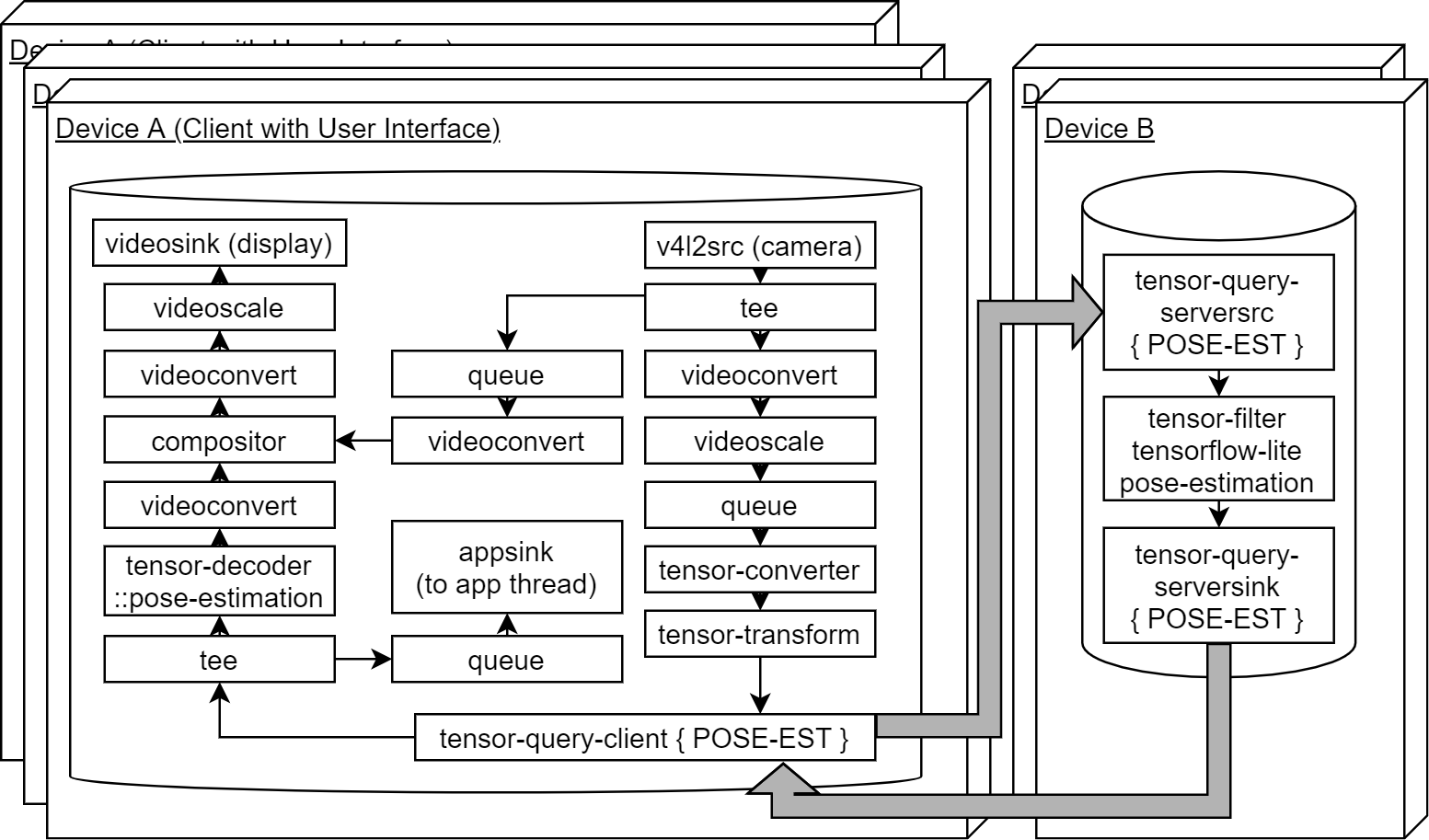}
    \caption{An offloading example with the proposed work.}
    \label{fig:SIMPLEOFFLOAD_NEW}
    \Description[A pipeline offloading workload to another pipeline via the proposed mechanism, query]{A pipeline of a client device with a user interface is offloading a pose-estimation inference to a pipeline running in another device with tensor-query client and server plugins. The client device has a single filter, tensor-query-client, that sends requests to and receives answers from a server. There can be multiple clients and multiple servers for a given topic, pose-estimation.}
\end{figure}

For R1 and R4, we propose query capability as pipeline filters, shown in Figure~\ref{fig:SIMPLEOFFLOAD_NEW}.
Client pipelines become trivial, and we can easily support multiple clients and multiple servers for an instance of service.
We integrate network protocols and serialization mechanisms in the client/server filters for the simplicity of pipelines.
For the robustness of services, it is desirable to allow multiple servers (Device B) compatible with given service requests from clients (Device A).

Another demonstrative among-device AI user scenario is a system with multiple input devices and independent processing and output devices.
The example application for this scenario has two input devices with USB cameras, one processing device (demonstrated with a Google Coral USB accelerator), and one output device (demonstrated with an LCD).
This scenario stands for home IoT systems with lightweight input devices (sensors, cameras, and microphones) and output devices (speakers and displays).
Vendors may expose their AI services via such lightweight devices (e.g., refrigerators, washers, and inexpensive display devices) while executing AI services at home for privacy protection, network latency reduction, and cloud cost reduction.
Such AI services can be consistently (for both latency and quality) provided across different user interfacing devices regardless of their processing power if they satisfy R1, R2, and R3.
In other words, such services can achieve the above if devices can discover available resources and services, connect to the discovered ones efficiently, and provide their resources to other devices.
If a vendor wants to expand their device ecosystem by inviting other vendors' devices, R5 and R6 are also required.

Some internal clients have specified a data serialization method for inter-pipeline connections: schemaless FlexBuffers of FlatBuffers.
We do not recommend using schemaless FlexBuffers for connecting stream pipelines; we recommend dynamic (flexible) schema instead.
Schemaless protocol usually incurs more overheads and run-time issues; we cannot verify data types at launch.
Moreover, being schemaless is meaningless in many among-device AI systems; i.e., it transfers the responsibility of interpreting the output of a neural network in a sender from the sender to the receiver.
In other words, except for not generating errors for data type mismatches (we do not recommend this but have failed to persuade the clients), it burdens the receiver to be more aware of the specifications of the sender.
However, we have accepted it with the condition of using FlexBuffers only for research prototypes and using ``other/tensors'' MIME types for products.
Note that having different data types between prototypes and products may incur unnecessary technical debts.

\begin{figure}
    \centering
    \includegraphics[width=0.98\linewidth]{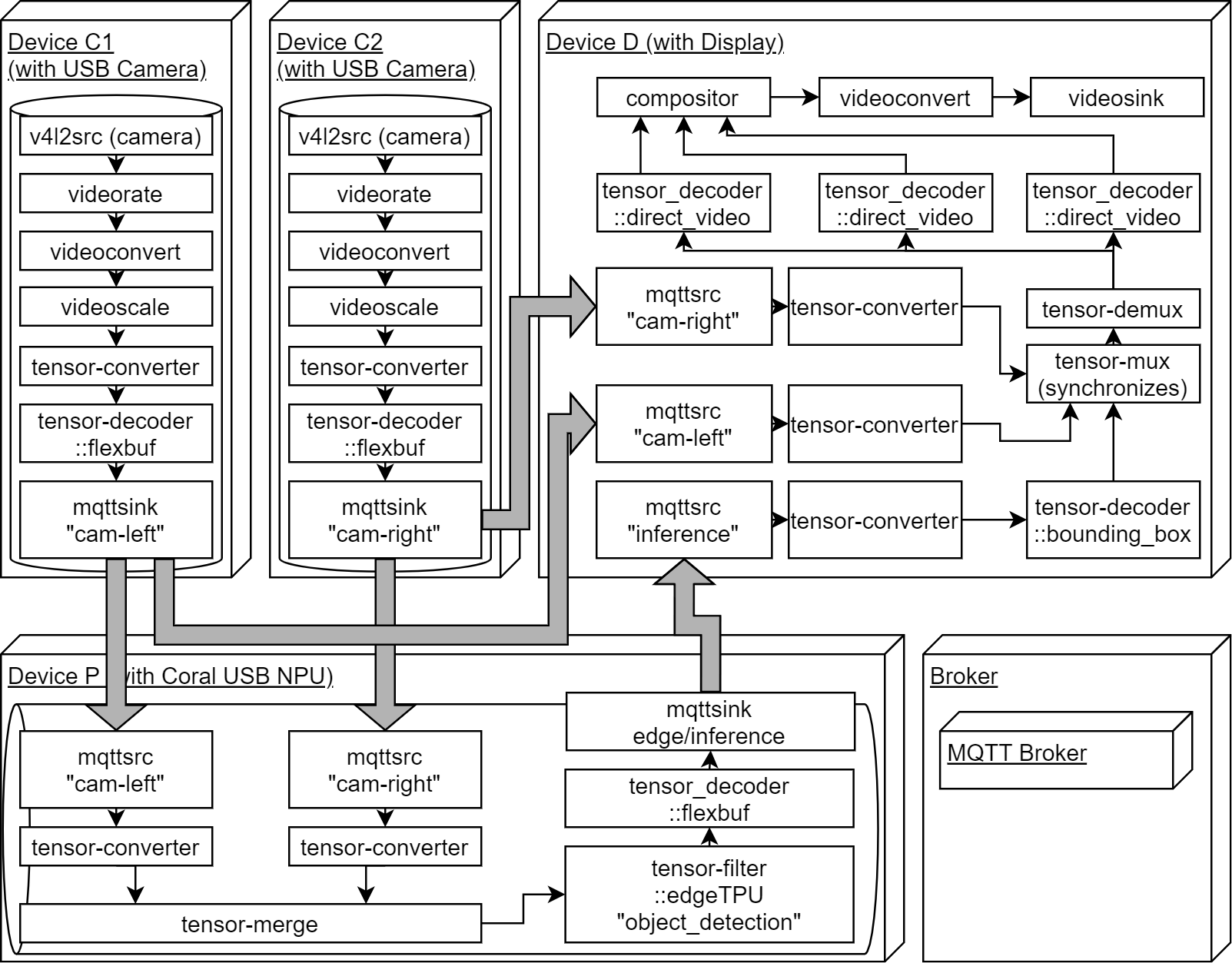}
    \caption{Stream pub/sub connections for distributed IoT AI application example.}
    \label{fig:PUBSUB_ORG}
    \Description[Pipelines connected via pub/sub streams]{Two pipelines are publishing live video from cameras, one processing pipeline that subscribes to the two camera streams and publishes inference results, and a displaying pipeline that subscribes to the other three published streams. Each pipeline runs on another device. An MQTT broker exists in another device as well.}
\end{figure}

Writing pipelines with raw network protocol plugins (TCP, UDP, and RTSP) over-complicates pipelines, as we have experienced with the previous example.
For example, we may have multiple pipelines sharing the output of a single pipeline.
We may also have a pipeline that needs to connect to any compatible pipeline available or does not want to specify network addresses.
Thus, after the initial prototype with ZeroMQ~\cite{ZMQ}, we have implemented MQTT~\cite{MQTT} stream pub/sub plugins compatible with general GStreamer data streams.
With MQTT pub/sub plugins, we have implemented prototype pipelines described in Figure~\ref{fig:PUBSUB_ORG}, showing that NNStreamer meets the requirements of the given application scenario.
Because MQTT requires a broker for relaying messages and discovering published services, users need to deploy an MQTT broker service.
In the figure, we have omitted flows of messages redirected by the broker.

With this prototype instance, the client's initial requirements (AI application developers) have been satisfied.
However, as it has always been, the client has found additional requirements after seeing the demonstration.
The client has requested another demonstration of pipelines that minimizes the difference between timestamp values of the two camera input frames when Device P merges frames from the two.
In this system, Device C1 and C2 mark timestamps when the data frame is created in the pipeline or given by the camera hardware.
Another additional requirement is the capability to compress data transmitted between devices.
The last additional requirement is the robustness of connections where alternative resources can be connected if one fails in run-time.

The first additional requirement is satisfied with the same pipeline architecture by updating MQTT plugins (mqttsink and mqttsrc) with additional features, including the NTP~\cite{NTP} protocol, which we discuss in the next section.
The second is about performance optimization, which is more appropriate with products than with prototypes.
By abandoning FlexBuffers from the pipelines, as mentioned above for products, we can easily apply compression mechanisms (zlib-gst~\cite{GSTZLIB}, JPEG for each frame, or MPEG real-time compression).
We further provide performance optimization for inter-pipeline transmissions with an additional protocol for MQTT, which we discuss in the next section.
The nature of MQTT and the way we use MQTT for inter-pipeline connections satisfy the last additional requirement: connect with the capability of pipelines, not with the address of devices.
Note that some clients have explicitly requested sparse tensor streams to compress streams for language and speech models.

\section{Implementation}\label{section:Implementation}

This section describes how we have upgraded NNStreamer for among-device AI based on the design described in the previous section: stream data types, protocols for inter-pipeline transmissions, and ``NNStreamer-Edge'' library for further compatibility with different vendors and operating systems.
Then, we describe how the requirements, R1 to R7, are met with the given implementation.
Lastly, we describe additional features assisting among-device AI features and updates from the previous work.

\subsection{Data types}

R2 does not only imply using FlexBuffers~\cite{Flexbuf} for inter-pipeline transmissions.
Data streams in a pipeline are often required to be schemaless (or dynamic schema).
For example, a pipeline may pre-process a live video stream with an object detection neural network  to crop the video, making a video stream filled with a detected human body.
Then, the pre-processed tensor stream representing the cropped video may have varying dimensions per frame, which another neural network (e.g., pose estimation) may use as its input stream.
Thus, even while schemaless data exchanges between pipelines might be useless, there is a need for the dynamic schema.

We update the tensor stream data type, ``other/tensors'' (MIME for GStreamer capability), so that users may specify the format of \textit{static}, \textit{dynamic}, and \textit{sparse}.
The conventional tensor stream with schema is \textit{static}, and it is the default format.
With \textit{dynamic} format, the dimension and type may vary for each frame in a stream, which can serve schemaless data from remote pipelines and stream with dynamic schema.
Unlike \textit{static} format, which does not have format information in the buffers of a data frame, the \textit{dynamic} format has a header in each data frame buffer that specifies dimension and type for each data frame.
The other format, \textit{sparse}, allows expressing tensors with the coordinate list format (COO)~\cite{COO}.

Filters for tensor streams, \textit{tensor\_*} elements, are updated to handle \textit{static} and \textit{dynamic} formats.
However, \textit{tensor\_*} elements do not directly handle \textit{sparse} formats because its binary representation of tensor data is not compatible with the other two formats.
Thus, we provide converting filters, \textit{tensor\_sparse\_enc} and \textit{tensor\_sparse\_dec}.

In addition to the Protocol Buffers and FlatBuffers support of the previous work, we add FlexBuffers of FlatBuffers for schemaless data transmissions as sub-plugins of \textit{tensor\_converter} (Flex\-Buffers to ``other/tensors'') and \textit{tensor\_decoder} (``other/tensors'' to Flex\-Buffers).

\subsection{Protocols}

There are various raw network protocol stream filters as off-the-shelf shared libraries of GStreamer.
According to GStreamer's reference page at \url{https://gstreamer.freedesktop.org}, they include TCP, UDP, RTSP, HTTP, HTTPS, FTP, HLS, and many others with different serialization mechanisms.
However, as explained in the previous section, they are not enough for among-device AI applications while they appear almost complete for multimedia applications and services.

\subsubsection{Pub/Sub Protocol}

Publish-subscribe architecture has been widely accepted by the robotics community, including ROS~\cite{ROS}.
AI application and service developers have requested a similar capability to publish services and subscribe to the services.
After iterations of different implementations from ROS, ZeroMQ~\cite{ZMQ}, and MQTT~\cite{MQTT}, we have decided to use MQTT as the basis.
ROS is not chosen because it is over-complicated for the need, requires recompiling software for updated schema, and excessively requires software packages for its dependencies.
On the other hand, ZeroMQ and MQTT are lightweight and do not require additional software packages for their dependencies.
ZeroMQ is the most lightweight among these; however, we have chosen MQTT because major home IoT standards (Matter~\cite{Matter} and SmartThings~\cite{SmartThings}) use MQTT already.
We want to make the among-device AI capability of NNStreamer compatible with such home IoT standards.

For pub/sub capability (Figure~\ref{fig:PUBSUB_ORG}), we provide two GStreamer plugins: \textit{mqtt\-sink} and \textit{mqtt\-src}.
With \textit{mqttsink}, a pipeline may publish a stream (output of the pipeline) declared with a topic string.
With \textit{mqttsrc}, a pipeline may subscribe to a published stream and fetch an input stream for the pipeline from an output stream of another pipeline discovered by the given topic string.
The topic string is independent of GStreamer capability (GSTCAP), representing the type information of a stream between two pipeline elements.
A schemaless stream may have a GSTCAP of ``other/flexbuf'', and a receiving pipeline should interpret as a properly structured stream type.
A dynamic-schema stream may have a GSTCAP of ``other/tensors,format=flexible'', and a receiving element (a neural network or a tensor processing element) can directly handle without additional conversion or interpretation.

We use ``Eclipse Paho MQTT C client library''~\cite{paho_mqtt_c} for MQTT implementation.
It is open source software with high coverage of MQTT features and high portability (an independent C/C++ library), which has a port for lightweight RTOS devices (e.g., FreeRTOS).
Via MQTT connections, among-device AI transport plugins transmit data and metadata, including the corresponding GSTCAP and data size.

\subsubsection{Query Protocol}\label{IMPL:Query}

We provide three GStreamer plugins for query capability (inference workload offloading in Figure~\ref{fig:SIMPLEOFFLOAD_NEW}): \textit{tensor\_query\_client}, \textit{tensor\_query\_server\-src}, and \textit{tensor\_query\_server\-sink}.
In a pipeline, \textit{tensor\_query\_client} behaves equivalently to \textit{tensor\_filter} representing a neural network model.
Thus, for other parts of the pipeline or the application running the pipeline (the client), \textit{tensor\_query\_client} hides all the details for inference task offloading transparently and may be switched with local on-device AI elements.
It sends queries (input stream for a neural network) to \textit{tensor\_query\_serversrc} and receives inference results from \textit{tensor\_query\_serversink}.
In a server-side pipeline, the two elements, \textit{tensor\_query\_serversrc} (input for the server) and \textit{tensor\_query\_serversink} (output for the server), are paired and share information of client connections.
In case there are multiple clients for a server-side pipeline, \textit{tensor\_query\_serversrc} tags a client ID to the stream's metadata, which \textit{tensor\_query\_serversink} accesses to choose the proper client connections.

Query elements implement two different transport protocols that users may choose: TCP-raw and MQTT-hybrid.
With TCP-raw, the connection between clients and servers are raw TCP connections, which does not provide the flexibility and robustness required by R3 and R4.
MQTT-hybrid transmits the connection and control information via MQTT connections, which easily satisfies R3 and R4 by allowing multiple server pipelines compatible with a given topic requested by a client.
To allow multiple servers for a given topic of a client pipeline, we use wildcards and topic filters for MQTT topics, which subscribers (clients) use to choose publishers (servers) dynamically.
For example, with servers of ``/objdetect/mobilev3'' and ``/objdetect/yolov2'', a client may choose either of them by subscribing to ``/objdetect/\#''.

MQTT-hybrid transmits data transmission via direct TCP connections without brokers for higher throughput and lower broker overheads.
Note that MQTT connections are not suitable for high-bandwidth streaming because of excessive overheads of a broker; thus, per clients' requests, we have designed the MQTT-Hybrid protocol for queries.
The clients of pub/sub have not yet requested higher bandwidth; however, we will provide MQTT-hybrid along with pure MQTT for pub/sub with the subsequent releases of NNStreamer.

\subsubsection{Timestamp Synchronization}\label{subsubsection:timestampsynchronization}

\begin{figure}
    \centering
    \includegraphics[width=0.98\linewidth]{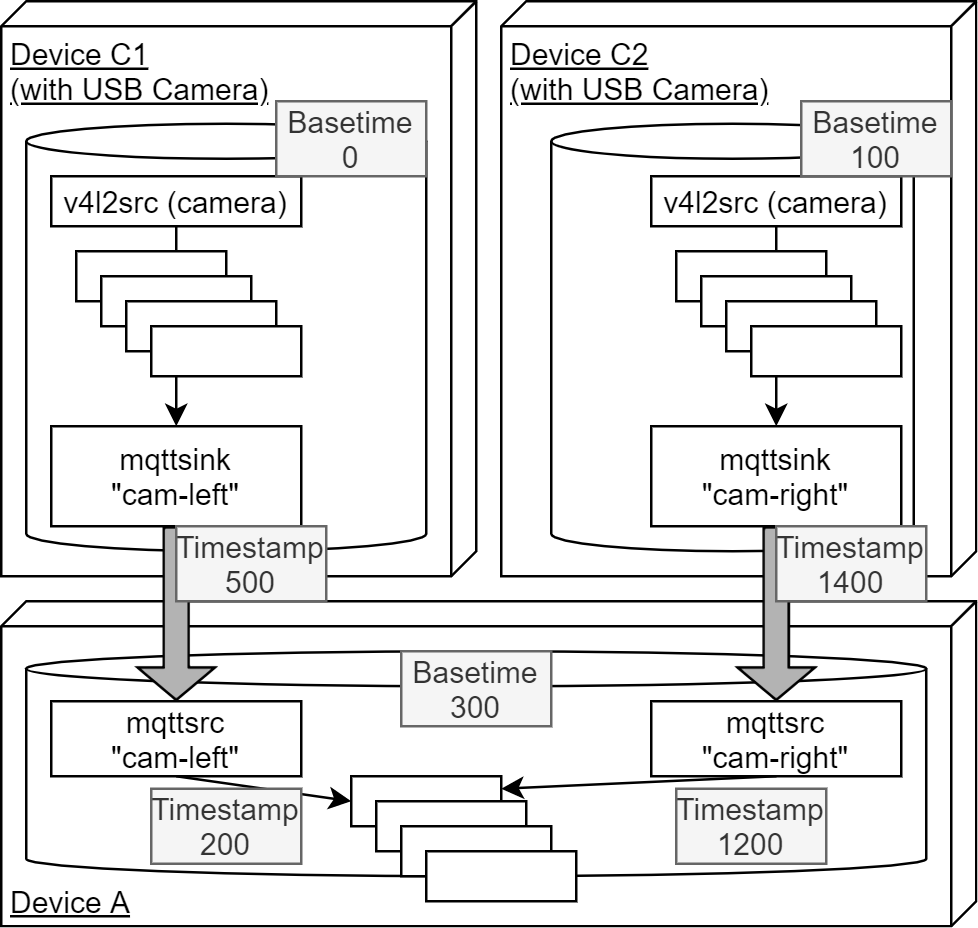}
    \caption{Timestamps from multiple source pipelines~\cite{nnstreamer_mqtt_sync_doc}}
    \label{fig:PUBSUB_TIMESTAMP}
    \Description[Three pipelines with different timing, but synchronized]{Three pipelines are running in different devices with different clocks. With the different base-time values of each pipeline transmitted to a subscribing pipeline, the subscribing pipeline updates timestamp values of incoming frames from publishing pipelines with its base-time to synchronize multiple input streams from different devices.}
\end{figure}

An internal client has requested to minimize temporal differences between multiple input sources.
For such needs, we add an inter-pipeline synchronization mechanism for pub/sub streams~\cite{nnstreamer_mqtt_sync_doc}.
Note that ``query'' does not suffer from such synchronization issues because client pipelines do not depend on the timestamp values of servers.
This timestamp synchronization mechanism makes stream publishers send base-time values of the publishing pipeline converted to universal time and relative timestamp values of buffers.
Then, receivers (subscribers) correct timestamp values of incoming buffers with their base-time.
The mechanism relies on clock synchronization between connected pipelines implemented by NTP protocol synchronizing relative clocks of inter-connected pipeline elements (\textit{mqttsink} and \textit{mqttsrc}).
We can test this by injecting latency to a publisher in Figure~\ref{fig:PUBSUB_ORG} by inserting the \textit{queue2} GStreamer plugin, which holds data until the specified size or duration.

\subsection{NNStreamer-Edge Library}

NNStreamer-Edge is a lightweight and portable library implementing the proposed protocols to connect with NNStreamer pipelines without adopting NNStreamer.
NNStreamer-Edge is an open source software package independent from NNStreamer and its basis, GStreamer: \url{https://github.com/nnstreamer/nnstreamer-edge}.
It does not depend on NNStreamer or GStreamer so that devices that cannot afford GStreamer or heavy operating systems may easily use NNStreamer-Edge.
NNStreamer-Edge has minimal library dependency, Paho MQTT C client library.
This library also has minimal library dependency; if SSL is disabled, it only requires essential tools (e.g., GCC and libc).

NNStreamer-Edge of October 2021 supports tensor stream publish for remote sensors and cameras: ``edge\_sensor'' module.
This module behaves like an ``mqttsink'' NNStreamer element with ``other/tensors'' streams to connect with ``mqttsrc'' NNStreamer elements.
We have designed other modules of NNStreamer-Edge, ``edge\_\-output'' and ``edge\_\-query\_\ client'', but have not yet released them.

The primary objective is to extend connectivity for devices of arbitrary vendors; thus, the proposed mechanisms should be compatible with or included in alliances such as Matter~\cite{Matter} and SmartThings~\cite{SmartThings}.
Because NNStreamer-Edge has chosen Apache 2.0 license and does not have extra dependencies, anyone may implement their proprietary software with NNStreamer-Edge.
For example, third-party developers may implement a proprietary MediaPipe plugin with NNStreamer-Edge so that arbitrary MediaPipe pipelines may communicate with NNStreamer pipelines.
Note that DeepStream pipelines are GStreamer pipelines; thus, they can trivially connect to NNStreamer pipelines, which are also GStreamer pipelines.

\subsection{How the requirements are met}

This section describes how the proposed implementation satisfies each requirement mentioned in the first section.

\begin{enumerate}[label=R\arabic*.]
    \item As long as we implement AI services and shared input/output streams as NNStreamer pipelines, we can deploy them atomically as pipeline instances or pipeline descriptions.
    With \textit{tensor\_query\_client/server}, we can offload inference tasks to another pipeline.
    With \textit{mqttsrc/sink}, a pipeline may serve as a data stream publisher or subscriber.
    \item If a stream is typed as ``other/flexbuf'' using ``Flexbuf'' sub-plugins, it is schemaless.
    A NNStreamer-native stream can be schemaless if it is typed as ``other/tensor,format=flexible''.
    The former is for compatibility with third-party software, and the latter is for in-pipeline or inter-pipeline streams.
    \item The adoption of MQTT for connections has satisfied this requirement.
    Multiple subscribers or clients can connect with a server with topic names.
    Multiple publishers or servers may be available for a given client or subscriber to choose with topic filters and wildcards.
    The timestamp synchronization mechanism described in Section~\ref{subsubsection:timestampsynchronization} supports synchronization.
    Sparse tensors and \textit{gst-gz}~\cite{GSTZLIB} support compressed transmissions.
    \item MQTT client libraries and MQTT brokers handle this.
    \item NNStreamer is LGPL 2.1 with all source codes included as default plugins to satisfy this requirement. For systems without NNStreamer and GStreamer, but with their proprietary middleware, we release NNStreamer-Edge with Apache 2.0 so that users may distribute their software without the condition of LGPL 2.1.
    Anyone may participate in designing and developing both packages in open space, Github.com.
    \item NNStreamer is cross-platform and deployed for different operating systems officially: Tizen, Android, Ubuntu, Yocto, and macOS.
    In addition, users have reported using it in Debian and OpenSUSE as well.
    It would not be too difficult to port it for Windows or iOS, but we have not tried it yet.
    The inter-pipeline connectivity library, NNStreamer-Edge, is designed to make the among-device AI capability compatible with different operating systems and different pipeline frameworks.
    \item We satisfy this by allowing developers to implement AI services with conventional NNStreamer plugins and adding inter-pipeline transmission plugins.
    For example, in Tizen, adding among-device AI capability (Tizen 6.5 M2) does not break the backward compatibility of the machine learning API set.
\end{enumerate}   
\section{Usage Example and Evaluation}

This section describes exemplar among-device AI systems with implementation details and experimental results.

\subsection{Workload Offloading with Query}

\begin{lstlisting}[caption={The code implementing workload offloading with query elements, depicted in Figure~\ref{fig:SIMPLEOFFLOAD_NEW}.}, label={lst:SIMPLEOFFLOAD_NEW}]
# Device A code
v4l2src ! tee name=ts
ts. videoconvert ! videoscale ! 
  video/x-raw,width=300,height=300,format=RGB !
  queue leaky=2 ! tensor_converter !
  tensor_transform ${TROPT}$ !
  tensor_query_client operation=${SVCNAME}$ !
  tee name=tc
ts. queue leaky=2 ! videoconvert ! mix.sink_1
tc. queue leaky=2 ! appsink name=appthread
tc. tensor_decoder mode=${DECMODE} ${DECOPTS} !
  videoconvert ! mix.sink_0
compositor name=mix sink_0::zorder=2 sink_1::zorder=1 ! videoconvert ! videoscale !
  video/x-raw,width=640,height=480 ! ximagesink
        
# Device B code
tensor_query_serversrc operation=${SVCNAME} !
tensor_filter framework=tensorflow-lite model=${MODELPATH}$ ! tensor_query_serversink
\end{lstlisting}

Listing~\ref{lst:SIMPLEOFFLOAD_NEW} shows a script code representing GStreamer pipelines in Figure~\ref{fig:SIMPLEOFFLOAD_NEW}.
GStreamer API or its wrapper, NNStreamer API (Tizen ML API), can execute such a script.
We can also execute the script directly on a shell with ``gst-launch'' for prototyping and testing.
We can apply such pipelines to home IoT systems consisting of a device with comfortable user interfaces without high computing power (e.g., an inexpensive TV) and a device with less comfortable user interfaces with high computing power (e.g., a home IoT hub or a mobile phone).
In other words, if a user has an inexpensive but large display (Device A) and a high-end mobile phone (Device B).
The user may run home fitness and training applications by offloading a pose-estimation neural network to Device B while using the camera and screen of Device A.

Please note the simplicity of server (Device B) code; declaring the service name (\textit{\$\{SVCNAME\}}) is all developers need to do.
If users want more operations in servers--e.g., pre-processing or data collecting--, they may add corresponding filters to the pipeline.
The client-side pipeline is straightforward as well.
The only changes from its on-device AI equivalent pipeline are replacing \textit{tensor\_filter} with \textit{tensor\_query\_client} and declaring the service name.
Then, we may have multiple clients and multiple servers for the given service name for resource sharing and robustness.

The string variables in Listing~\ref{lst:SIMPLEOFFLOAD_NEW} depend on neural network models and their input and output formats.
For example, if it is a Mobilenet V2 object detection model~\cite{MobileSSDv2} from TensorFlow Hub with COCO 2017 dataset~\cite{MobileSSDv2COCO}, the variables are:
\begin{lstlisting}
TROPT = "mode=arithmetic option=typecast:float32,add:-127.5,div:127.5"
SVCNAME = "objectdetection/ssdv2"
DECMODE = "bounding_boxes"
DECOPTS = "option1=mobilenet-ssd option2=/PATH/coco_labels_list.txt option3=/PATH/box_priors.txt option4=640:480 option5=300:300"
MODELPATH = /PATH/ssd_mobilenet_v2_coco.tflite
\end{lstlisting}

Configurations and behaviors of queues and merging points (compositor in this example) are crucial for the efficiency of parallelism.
With the \textit{leaky=2} option, a \textit{queue} drops older buffers if it becomes full.
Users may alter options, including the size of the queue and leaky modes, for further optimization.

\subsection{Stream Pub/Sub}

\begin{lstlisting}[caption={The code implementing remote sensors with pub/sub in Figure~\ref{fig:PUBSUB_ORG} along with timestamps mechanisms.}, label={lst:remotesensors}]
# Device C1 / C2
v4l2src do-timestamp=true ! videoconvert !
videorate ! video/x-raw,width=${W},height=${H},format=RGB,framerate=10/1 !
tensor_converter ! tensor_decoder mode=flexbuf !
queue ${OPT} ! mqttsink pub-topic=${CAM} sync=true

# Device P
tensor_merge name=m mode=linear option=1 !
tensor_decoder mode=direct_video ! videoscale !
video/x-raw,width=300,height=300,format=RGB !
tensor_converter ! tensor_filter framework=edgetpu, model=${MODELPATH} !
tensor_decoder mode=flexbuf !
mqttsink pub-topic=edge/inference

mqttsrc sub-topic=camleft is-live=false !
other/flexbuf ! tensor_converter ! queue !
m.sink_0
mqttsrc sub-topic=camright is-live=false !
other/flexbuf ! tensor_converter ! queue !
m.sink_1

# Device D
compositor name=mix sink_0::xpos=1 sink_0::ypos=0 sink_0::zorder=0 sink_1::xpos=${W} sink_1::ypos=0 sink_1::zorder=0 sink_2::xpos=1 sink_2::ypos=0 sink_2::zorder=1 !
videoconvert ! ximagesink

tensor_mux name=mux ! queue !
tensor_demux name=dmux

dmux.src_0 ! tensor_decoder mode=direct_video !
queue ! mix.sink_0
dmux.src_1 ! tensor_decoder mode=direct_video !
queue ! mix.sink_1
dmux.src_2 !
tensor_decoder mode=direct_video option1=RGBA !
queue ! mix.sink_2

mqttsrc sub-topic=camleft is-live=false !
other/flexbuf ! tensor_converter ! queue !
mux.sink_0
mqttsrc sub-topic=camright is-live=false !
other/flexbuf ! tensor_converter ! queue !
mux.sink_1

mqttsrc sub-topic=edge/inference is-live=false !
queue ! other/flexbuf ! tensor_converter !
other/tensors,num_tensors=4,dimensions="4:20:1:1,20:1:1:1,20:1:1:1,1:1:1:1",types="float32,float32,float32,float32" !
  option1=tf-ssd option2=${LABELPATH}
  option3=0:1:2:3,40 option4=${W}:${H}
  option5=300:300 !
tensor_converter ! queue ! mux.sink_2
\end{lstlisting}

Listing~\ref{lst:remotesensors} shows a GStreamer pipeline description of the application example shown in Figure~\ref{fig:PUBSUB_ORG}, where FlexBuffers~\cite{Flexbuf} serializes published streams.
When the \textit{v4l2src} creates a video frame buffer, fetching video streams from a USB camera attached to a Raspberry Pi 4 board in our experiments, the \textit{v4l2src} provides a timestamp value with the option of \textit{do-timestamp=true} to enable the mechanism in Figure~\ref{fig:PUBSUB_TIMESTAMP}.
Note that this code does not require the camera to be a USB camera or the system to be a Raspberry Pi 4.
The same code works for an embedded camera of a mobile phone or a typical Linux desktop PC without modification.
If multiple cameras are attached to the system, users may specify it with additional options; otherwise, it uses \textit{/dev/video0}.

The pipeline topology of the target application in Figure~\ref{fig:PUBSUB_ORG} has become more complex than the application in Figure~\ref{fig:SIMPLEOFFLOAD_NEW}.
It would require a considerable amount of time and effort to implement such systems without pipeline frameworks as we have experimented in~\cite{NNStreamer}; i.e., it would probably require well over thousands of lines of codes.
With the off-the-shelf GStreamer/NNStreamer plugins and effortlessly applied pipe-and-filter architecture, users can write such an among-device AI system within 100 lines of codes, which is easy to extend and port.
We can execute the same pipeline descriptions in Ubuntu PC, Yocto devices, Tizen devices (TVs, home appliances, robots, and wearable devices), or Android mobile phones.
With the NNStreamer-Edge library, developers can interconnect pipelines with more varying devices, including MediaPipe/DeepStream pipelines running in clouds or workstations and lightweight devices (microcontrollers) running RTOS.

Note that FlexBuffers, FlatBuffers, or Protocol Buffers are not required to connect NNStreamer pipelines and NNStreamer-Edge instances only.
For other independent non-NNStreamer processes subscribing to the published streams, we can apply FlatBuffers, FlexBuffers, or Protocol Buffers, popular data serialization mechanisms.

\subsection{Multi-device and Multi-modal}

\begin{figure}
    \centering
    \includegraphics[width=0.98\linewidth]{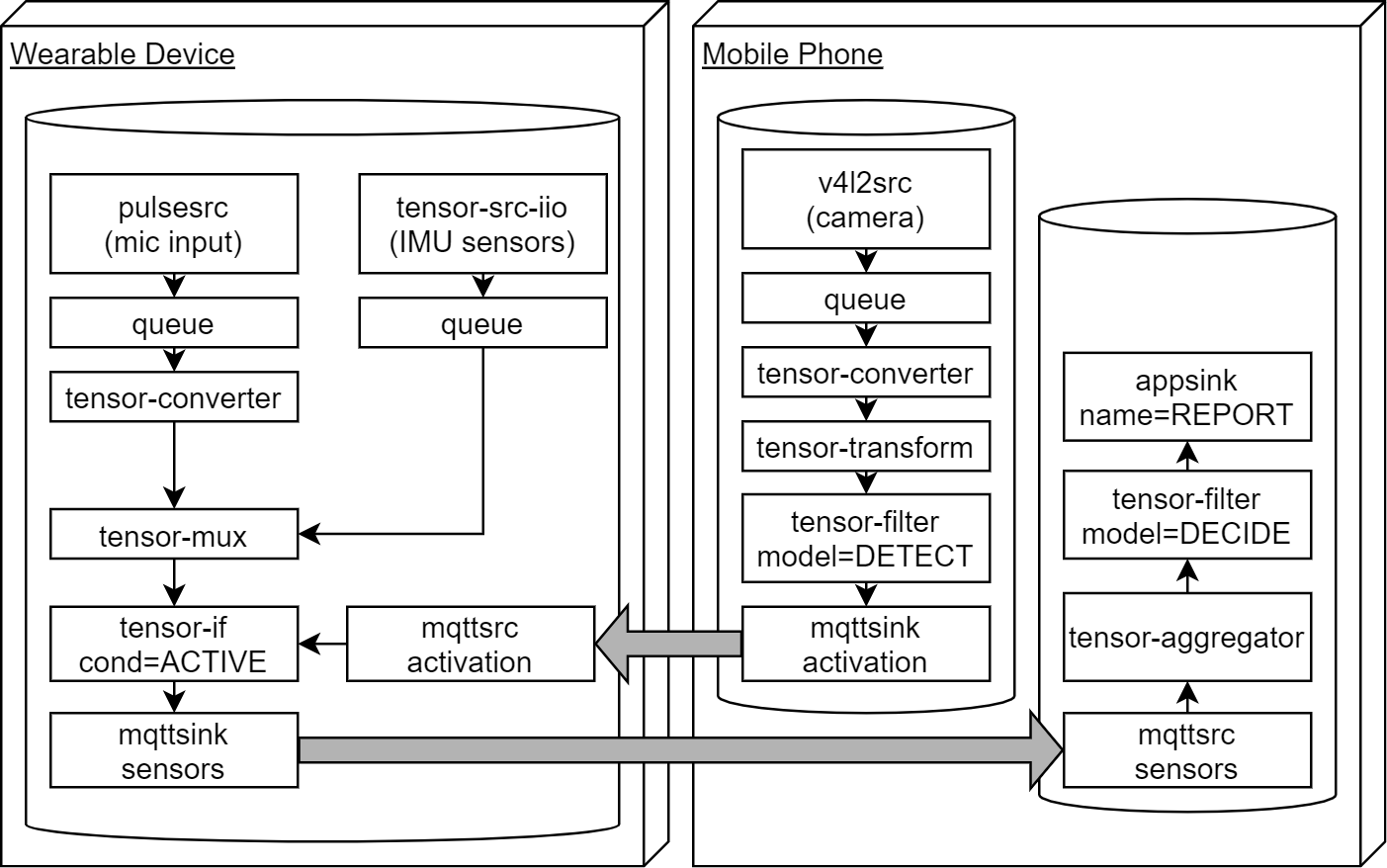}
    \caption{Augmented worker application with a wearable device and a mobile device.}
    \label{fig:AugmentedWorker}
    \Description[One pipeline in a wearable device, two pipelines in a mobile phone]{There is one pipeline in a wearable device that sends pre-processed sensor values to a pipeline of a mobile phone if another pipeline of the mobile phone says a data stream is required. There are two pipelines in a mobile device. The first one does inferences on a camera of the mobile phone to determine if data stream from the wearable device is required or not. The second one subscribes to the data stream of the wearable device and decides whether the worker has done a mistake or not with the incoming sensor values from the wearable device.}
\end{figure}

Another example, Figure~\ref{fig:AugmentedWorker}, shows an augmented worker application, which is both multi-device and multi-modal.
Such a system may assist workers in manufacturing plants by detecting and notifying events requiring attention.
For example, if a worker assembles parts incorrectly, the system sends an alarm to the worker.

In the left-hand side pipeline of the mobile device, the ``DETECT'' model detects if an action of assembling parts is starting and let the wearable device know.
Then, the wearable device will start streaming related data from the microphone and IMU sensors back to the mobile device.
Then, the right-hand side pipeline of the mobile device decides whether the assembling activity is correct or incorrect based on the data from the wearable device and reports to the application logic.
To further optimize power consumption in the wearable device, we may turn on and off the sensors based on the ``activation'' signal from the mobile device.
As we demonstrate with previous examples, the pipeline description code of this example incurs short lines of codes as well; i.e., usually a single line per block with exceptions of \textit{tensor\_if}, where we need to describe the condition with a script or a C function.  

\subsection{Performance Evaluation}

\begin{figure}
    \centering
    \includegraphics[width=0.99\linewidth]{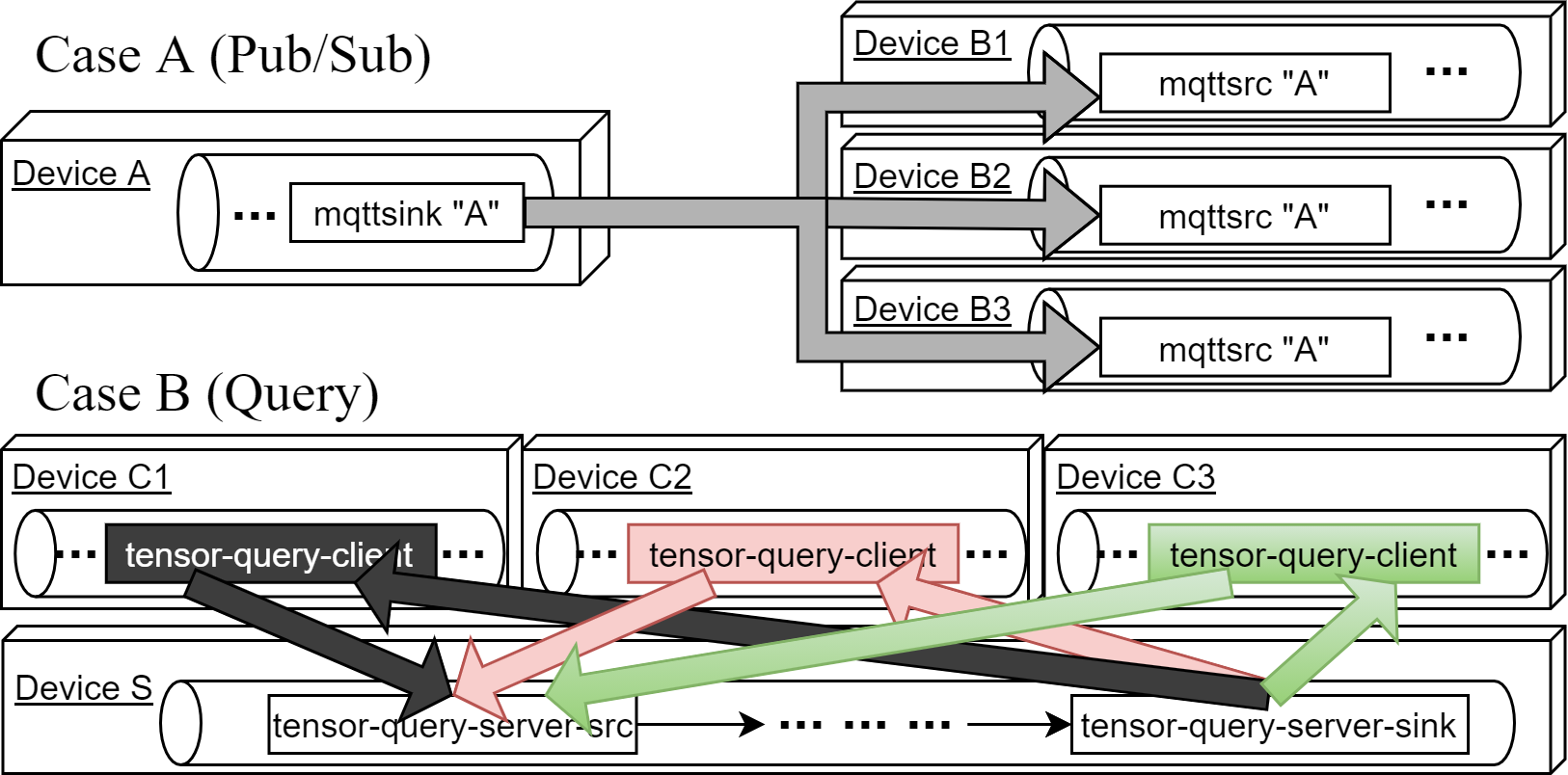}
    \caption{Pipelines for performance evaluation.}
    \label{fig:PerfExpTopology}
    \Description[Pipelines used for the evaulation]{There are two sets of pipelines shown. Case A set has one publishing pipeline and three subscribing pipelines. Case B set has one query server pipeline and three query client pipelines.}
\end{figure}

We evaluate the proposed among-device AI transport mechanisms: MQTT pub/sub and MQTT-hybrid query along with their lighter and faster counterparts: ZeroMQ pub/sub and TCP query.
We have experimented with Raspberry Pi 4 boards connected via Ethernet and NNStreamer 2.1.0-unstable.
Figure~\ref{fig:PerfExpTopology} shows the evaluated among-device AI pipelines.
Case A evaluates the stream pub/sub of MQTT and ZeroMQ.
Case B evaluates the query client/server streams of MQTT-hybrid and TCP direct connections.
We measure the throughput, CPU usage, and peak memory consumption to evaluate both performance and overhead.
We experiment with three different bandwidths of input streams from Device A and Device C: high, mid, and low bandwidths.
Each bandwidth corresponds to a Full-HD video stream, a VGA (640x480) video stream, and a QQVGA (160x120) video stream with a 60 Hz framerate.

\begin{figure*}
    \centering
    \includegraphics[width=0.98\textwidth]{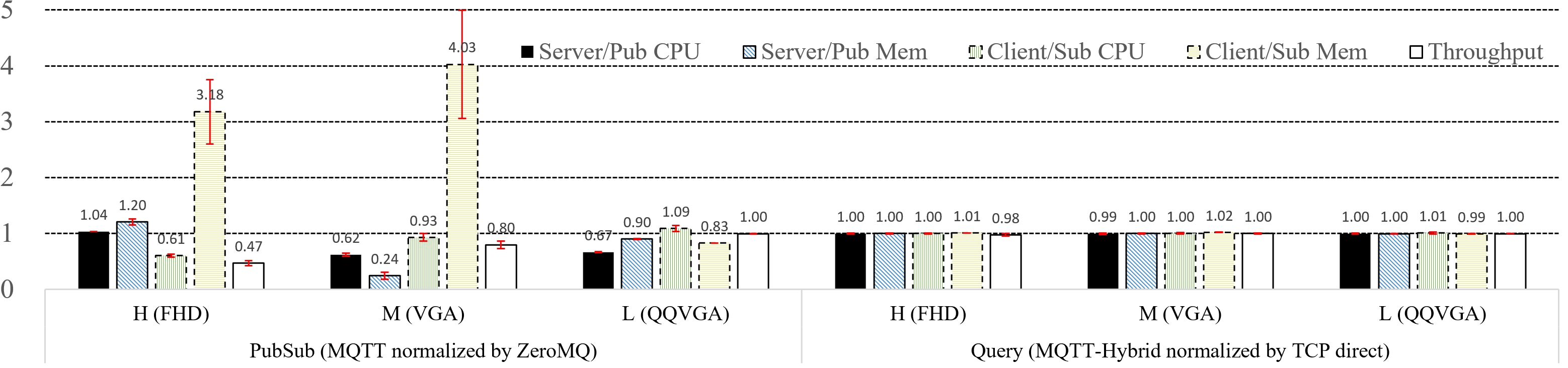}
    \caption{Evaluation results of Pub/Sub (MQTT normalized by ZeroMQ) and Query (MQTT-Hybrid normalized by TCP).}
    \label{fig:PerfExpResult}
    \Description[The graphs comparing performance]{The performance and overhead comparison. MQTT pub/sub is compared against ZeroMQ pub/sub. MQTT-hybrid query is compared against TCP query.}
\end{figure*}

Figure~\ref{fig:PerfExpResult} shows the evaluation results of MQTT pub/sub and MQTT-hybrid query, normalized by their counterparts, ZeroMQ and TCP, respectively.
Bars in the figure are normalized standard deviations, and values average five experimental runs of 30 seconds.
Case M and H have failed to transmit 60 Hz, implying that the network bottlenecks the throughput.
MQTT suffers from lower throughput and higher client memory overhead with high bandwidth streams (M and H).
MQTT-hybrid successfully eliminates the performance overheads of MQTT while keeping the rich features of MQTT.
For example, with the MQTT-hybrid query, if multiple servers are compatible with the client's topic, the client pipeline is automatically switched to another server when a connected server becomes unavailable.
Moreover, server pipelines may declare additional specifications for clients to choose, e.g., ``server workload status'' and ``neural network model and version''.
Note that the query plugins have both TCP direct and MQTT-hybrid implementations switchable at run-time.
For higher performance, we plan to add MQTT-hybrid for pub/sub plugins, as well.




\section{Conclusion}

We have extended the on-device AI pipeline framework~\cite{NNStreamer} for among-device AI systems (often marketed as edge AI).
The proposed framework, NNStreamer, is developed and released at the GitHub organization owned by Linux Foundation, \url{https://github.com/nnstreamer}, where anyone may participate in discussions and code contributions.
It is deployed to the mentioned operating systems daily or regularly: Tizen (as default machine learning framework), Android (for Android Studio via public repositories), Yocto (via meta-neural-network layer), Ubuntu (Launchpad PPA), macOS (homebrew repository).
We are deploying the among-device AI capabilities introduced in this paper via the Tizen 6.5 M2 release in October 2021; note that a few features (NNStreamer-Edge and MQTT-hybrid) are not yet approved by internal clients and are omitted in Tizen 6.5 releases but available in GitHub.
A few home appliances of 2022, including TVs, will be using Tizen 6.5 M2; thus, the mentioned features will be available for products in 2022, and Tizen Studio users writing applications for such products.

\subsection{Lessons Learned}

After wide deployment of the previous work~\cite{NNStreamer} across many prototypes and products, including mobile phones, wearable devices, TVs, and home appliances, we have worked on prototypes of various internal clients based on the proposed among-device AI methods. As a result, we have observed the following lessons and future work.

Many users appear to feel barriers against adopting pipe-and-filter architecture.
It is easy to show that pipelines with NNStreamer work appropriately and significantly better than the conventional implementation for both developmental costs and run-time performance.
However, we could have observed unexpectedly steep learning curves to adopt pipeline concepts and to describe pipeline topology.
For the former issue, we are preparing to write more diverse pipeline examples for users.
For the latter issue, we are implementing a WYSIWYG pipeline editor with a converter translating between GStreamer scripts and MediaPipe scripts (pbtxt) to re-use MediaPipe's pipeline editor.

Analyzing and profiling pipeline performance becomes more complicated with among-device AI pipelines.
We have an AI pipeline profiling tool, nnshark, which forks 
GstShark~\cite{GstShark}, created by a group of undergraduate students as an open source software project.
We have ported nnshark for Tizen and Android devices and deployed it to users.
However, with among-device AI capability, users are not satisfied with nnshark, and request profiling capability for the whole system consisting of multiple pipelines simultaneously.

We have been implementing initial prototypes or practicing pair programming for internal clients to address learning curves and developer relations.
For external users of open source communities, we try to catch up with technical questions in various channels (Slack, Github issues, mailing lists, and direct contacts) in addition to documents and sample applications.
However, with frequent requests from internal clients and our roadmap of new features, writing documents and sample applications have always had lower priority and been behind schedule.
Besides, with a few core contributors focusing on code contributions, it is challenging to motivate writing documents.
It is also difficult to justify having a dedicated document writer while we directly support internal clients so that they do not even bother to read the documents.

NNStreamer has external users that have contacted the authors: NXP Semiconductors, Collabora, ODKMedia, and fainders.ai.
Feedback and contributions from them are constructive for identifying issues and requirements.
Motivating external users to communicate with maintainers is essential; they are often too shy to do so.
We recommend keeping such communication in public channels to motivate other users to communicate with maintainers and the community.

Users often write pipelines incorrectly or inappropriately, and it is usually too late when we find it out.
For example, an NNStreamer application~\cite{HU.NNStreamer} drawing bounding boxes of detected objects with live video streams has an inappropriate pipeline design.
Creating video streams from neural network output is supposed to be handled by \textit{tensor\_decoder}.
Its authors~\cite{HU.NNStreamer} are aware of it; however, for neural network output formats not supported by default sub-plugins of \textit{tensor\_decoder}, they have abandoned \textit{tensor\_decoder} and made their application thread directly decode outputs and push decoded data into the pipeline.
It is supposed to write a \textit{tensor\_decoder} sub-plugin for such a case, which incurs lower overhead, higher throughput, and less implementation effort.
Fortunately, although inappropriately implemented, their NNStreamer implementation is shown to beat Open CV implementation with higher throughput.
Promoting communication with the NNStreamer community may address this issue; however, we have observed similar cases with internal users: the users have sometimes released inappropriate pipelines before addressing them.
Such releases are especially troubling considering that we are still at an early stage of deploying the pipeline paradigm to AI developers in the affiliation.
Besides, the standard practice of software verification and testing is not ready for such issues.
We may need another verification and testing process for newly introduced frameworks with a new concept, allowing framework developers to audit.
Such processes will provide more usage cases for framework developers and more chances to improve their frameworks.

\subsection{Directions of Evolution}

NNStreamer has started evolving for among-device AI systems.
Besides the apparent performance optimization, we have suggestions for among-device AI systems and pipeline frameworks as future work.

With frameworks including NNStreamer~\cite{NNStreamer}, DeepStream~\cite{deepstream}, and MediaPipe~\cite{mediapipe}, developers have started writing AI systems as pipelines.
These frameworks provide various documents and sample pipelines to mitigate difficulties suffered by developers writing AI pipelines.
However, as our users have complained, they are not enough.

We would like to suggest another method directly intervening pipeline development process.
First, we need to provide common parts of pipelines (sub-pipelines) as libraries; that developers can invoke or insert sub-pipelines in their pipelines.
There are various common parts in different AI applications: e.g., pre-processing video streams for object detection or audio streams for RNN-T~\cite{RNNT}.
This feature may often prevent inappropriately designing pipelines as well.

Then, we need a pipeline run-time repository where processes may register pre-defined pipelines, and other processes may invoke such pipelines.
This feature enables operating systems or middleware to register pipelines for applications.
It enables applications without particular AI features to invoke such pipelines without actually writing pipelines.
Moreover, suppose a vendor wants to separate the application development and AI development divisions; this approach will cleanly separate code repositories for corresponding divisions as a client has wanted.

We have further future directions that require more time and effort; the above can be implemented and deployed within a year.
We envision an IoT ecosystem where devices of any vendors may join, which is related to R5 and R6.
Matter~\cite{Matter} along with SmartThings~\cite{SmartThings} is a promising candidate for such an IoT ecosystem; however, it lacks a data transmission protocol for inter-device AI services.
In the future, we expect to propose such a protocol with among-device AI capabilities mentioned in this paper with extensions of interoperability with microcontrollers and other pipeline frameworks.
With such protocols included in the IoT ecosystem, connected devices will be able to consistently provide AI services regardless of interface devices' computation power.
This inclusiveness and consistency will promote the proliferation of various on-device and among-device AI services without the need for exposing privacy and private data to external computing nodes.

\begin{acks}
We greatly appreciate contributions from the open source software community with various affiliations: not only Samsung but also NXP Semiconductors, Collabora, universities, and independent hobbyists.
We also appreciate developers and users who gave us precious feedback and other core NNStreamer contributors, including Jijoong Moon, Parichay Kapoor, Dongju Chae, Jihoon Lee, and Hyeonseok Lee.
The discussions with and advice from the following Samsung Research members have especially been crucial for this project; Daehyun Kim, Gu-Yeon Wei, JeongHoon Park, Semun Lee, Jinmin Chung, Sunghyun Choi, Daniel Dongyuel Lee, and Sebastian Seung.
This project is supported by Samsung Research of Samsung Electronics and LF AI \& Data Foundation.
\end{acks}

\bibliographystyle{ACM-Reference-Format}
\bibliography{main}

\end{document}